\documentclass[letterpaper, 10 pt, conference]{ieeeconf}  

\IEEEoverridecommandlockouts                              
\usepackage{times}
\usepackage{epsfig}
\usepackage{graphicx}
\usepackage{amsmath}
\usepackage{amssymb} 
\usepackage{color}

\title{\LARGE \bf
DeepSIC: Deep Semantic Image Compression
}
\author{  Sihui Luo\\
Zhejiang University\\
Hangzhou, China\\
{\tt\small}
\and
Yezhou Yang\\
Arizona State University\\
\and 
Mingli Song\\
Zhejiang University \\
Hangzhou, China\\
}

\begin{document}

\maketitle
\thispagestyle{empty}
\pagestyle{empty}

\begin{abstract}
Incorporating semantic information into the codecs during image compression can significantly reduce the repetitive computation of fundamental semantic analysis (such as object recognition) in client-side applications. The same practice also enable the compressed code to carry the image semantic information during storage and transmission. In this paper, we propose a concept called Deep Semantic Image Compression (DeepSIC) and put forward two novel architectures that aim to reconstruct the compressed image and generate corresponding semantic representations at the same time. The first architecture performs semantic analysis in the encoding process by reserving a portion of the bits from the compressed code to store the semantic representations. The second performs semantic analysis in the decoding step with the feature maps that are embedded in the compressed code. In both architectures, the feature maps are shared by the compression and the semantic analytics modules. To validate our approaches, we conduct experiments on the publicly available benchmarking datasets and achieve promising results. We also provide a thorough analysis of the advantages and disadvantages of the proposed technique.
%
%
%
%
  
\end{abstract}

\section{Introduction}\label{sec:introduction}



As the era of smart cities and Internet of Things (IoT) unfolds, the increasing number of real-world applications require corresponding image and video transmission services to handle compression and semantic encoding at the same time, hitherto not addressed by the conventional systems. Traditionally, image compression process is merely a type of data compression that is applied to digital images to reduce the storage and transmission cost of them. Almost all the image compression algorithms stay at the stage of low-level image representation in which the representation is arrays of pixel values. For example, the most widely used compression methods compress images through pixel-level transformation and entropy encoding~\cite{Wallace1992The,franzen1999kodak}. However, these conventional methods do not consider encoding the semantic information (such as the object labels, attributes, and scenes), beyond low-level arrays of pixel values. At the same time, the semantic information is critical for high-level reasoning over the image.

Within the last decade, Deep Neural Networks (DNN) and Deep Learning (DL) have laid the foundation for going beyond hand-crafted features in visual recognition, with a significant performance boost in topics ranging from object recognition~\cite{krizhevsky2012imagenet}, scene recognition~\cite{Van2010Evaluating,NIPS2014_5349,Herranz2016Scene}, action recognition~\cite{Wang2011Action}, to image captioning~\cite{donahue2014long, karpathy2014deep, vinyals2014show,chen2014learning} and visual question answering~\cite{VQA,zhu2015visual7w,lu2016hierarchical}. Recently, several significant efforts of deep learning based image compression methods have been proposed to improve the compression performance~\cite{balle2016end,Toderici2015variable,Toderici2016full,gregor2016towards,Theis2017lossy,Rippel2017real}.

As Rippel and Bourdev~\cite{Rippel2017real} point out, generally speaking, image compression is highly related to the procedure of pattern recognition. In other words, if a system can discover the underlying structure of the input, it can eliminate the redundancy and represent the input more succinctly. Recent deep learning based compression approaches discover the structure of images by training a compression model and then convert it to binary code~\cite{balle2016end,Toderici2015variable,Toderici2016full,gregor2016towards,Theis2017lossy,Rippel2017real}. Nevertheless, to the best of our knowledge, a deep learning based image compression approach incorporating semantic representations has not yet been explored in the literature. These existing DL-based compression codecs, like the conventional codecs, also only compress the images at pixel level, and do not consider the semantics of them. Currently, when the client-side applications require the semantic information of an image, they have to firstly reconstruct the image from the codec and then conduct an additional computing to obtain the semantic information.

\emph{Can a system conduct joint-optimization of the objectives for both compression and the semantic analysis?} In this paper, we make the first attempt to approach this challenging task which stands between the computer vision and multimedia information processing fields, by introducing the Deep Semantic Image Compression~(DeepSIC). Here, our DeepSIC framework aims to encode the semantic information in the codecs, and thus significantly reduces the computational resources needed for the repetitive semantic analysis on the client side.
%
%
%
%

%
%
%
%

\begin{figure}[t]
\begin{center}
   \includegraphics[width=0.48\textwidth]{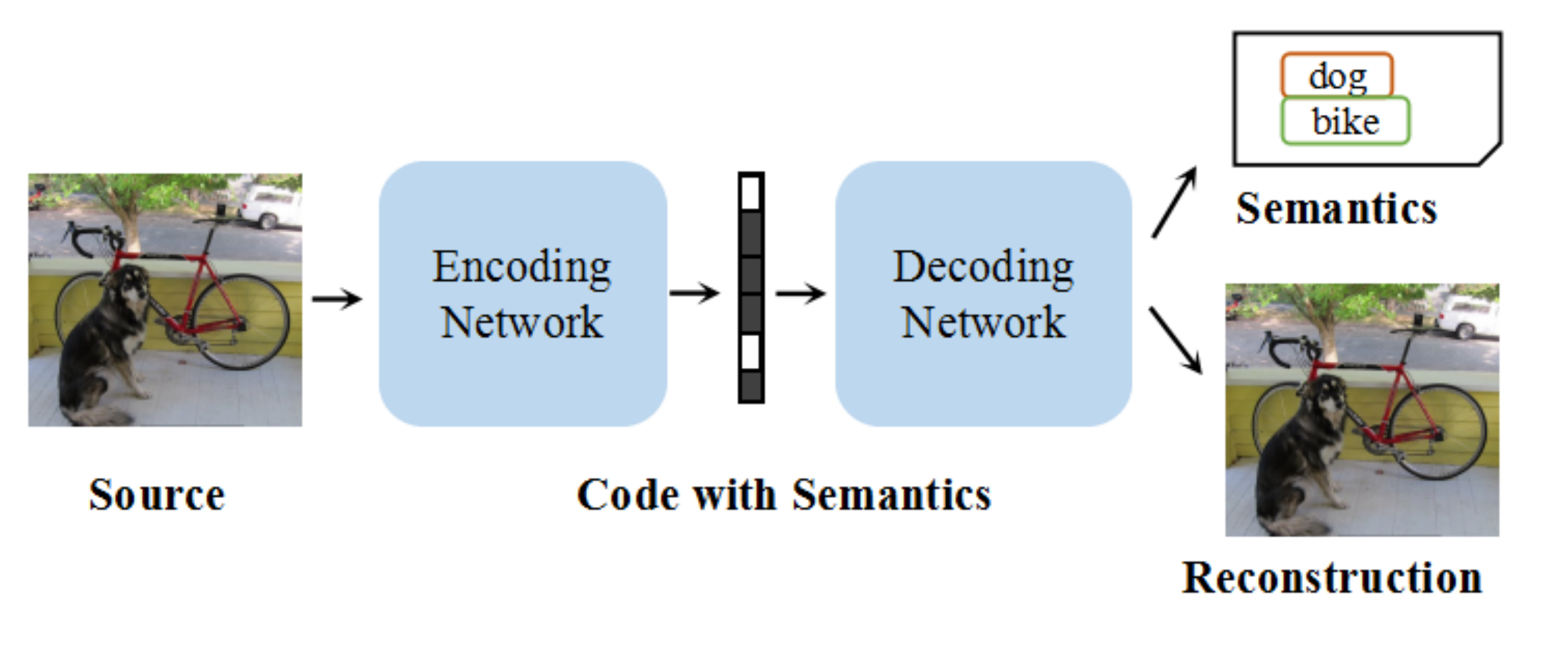}
\end{center}
   \caption{General semantic image compression framework}
\label{fig:framework}
\end{figure}

We depict the DeepSIC framework in Figure~\ref{fig:framework}, which aims to incorporate the semantic representation within the codec while maintaining the ability to reconstruct visually pleasing images. Two architectures of our proposed DeepSIC framework are given for the joint analysis of pixel information together with the semantic representations for lossy image compression. One is pre-semantic DeepSIC, which integrate the semantic analysis module into the encoder part and reserve several bits in the compressed code to represent the semantics. 
The other is post-semantic DeepSIC, which only encodes the image features during the encoding process and conducts the semantic analysis process during the reconstruction phase. The feature retained by decoding is further adopted in the semantic analysis module to achieve the semantic representation.
%
%
%
%
%
%
%
%

In summary, we make the following contributions: \begin{itemize}
\item We propose a concept called Deep Semantic Image Compression that aims to provide a novel scheme to compress and reconstruct both the visual and the semantic information in an image at the same time. To our best knowledge, this is the first work to incorporate semantics in image compression procedure.
\item We put forward two novel architectures of the proposed DeepSIC: pre-semantic DeepSIC and post-semantic DeepSIC.
\item We conduct experiments over multiple datasets to validate our framework and compare the two proposed architectures.
\end{itemize}

The rest of this paper is organized as follows: In section~\ref{sec2}, the conventional and DL-based image compression methods are briefly reviewed. Section~\ref{sec3} describes the proposed DeepSIC framework and the two architectures of DeepSIC in detail. Our experiment results and discussions are presented in section~\ref{sec4}. Finally, we conclude the paper and look into the future work in section~\ref{sec5}.

\section{Related work}\label{sec2}

Standard codecs such as JPEG~\cite{Wallace1992The} and JPEG2000~\cite{rabbani2002overview} compress images via a pipeline which roughly breaks down to 3 modules: transformation, quantization, and entropy encoding. It is the mainstream pipeline for lossy image compression codec. Although advances in the training of neural networks have helped improving performance in existing codecs of lossy compression, recent learning-based approaches still follow the same pipeline as well~\cite{balle2016end,Toderici2015variable,Toderici2016full,gregor2016towards,Theis2017lossy,Rippel2017real}. 

These approaches typically utilize the neural networks to retain the visual features of the image and then convert them into binary code through quantization. Some of these approaches may further compress the binary code via an entropy encoder. The feature extraction process replaces the transformation module in JPEG, which automatically discovers the structure of image instead of engineers manually do. Toderici~\cite{Toderici2015variable, Toderici2016full} explore various transformations for binary feature extraction based on different types of recurrent neural networks and compressed the binary representations with entropy encoding.~\cite{balle2016end} introduce a nonlinear transformation in their convolutional compression network for joint-optimization of rate and distortion, which effectively improves the visual quality.

Many recent works utilize autoencoders to reduce the dimensionality of images. Promising results of autoencoders have been achieved in converting an image to compressed code for retrieval~\cite{Krizhevsky2012Using}. Autoencoders also have the potential to address the increasing need for flexible lossy compression algorithms~\cite{Theis2017lossy}. Other methods expand the basic autoencoder structure and generate the binary representation of the image by quantizing the bottleneck layer.
 
More recently, since Goodfellow~\cite{Goodfellow2014Generative} introduced Generative Adversarial Networks(GANs) for image generation, GANs have been demonstrated to be promising in producing fine details of the images~\cite{johnson2016perceptual,im2016generating}. In the compression field, GANs are usually employed to generate reconstructed images that look natural, with the naturalness measured by a binary classifier. The intuition is, if it is hard to distinguish the generated images from the original, then the generated images are "natural" enough for humans. Some works have achieved significant progress in generating smaller compressed code size but more visually pleasing reconstructed images. Gregor et al.~\cite{gregor2016towards} introduce a homogeneous deep generative model in latent variable image modeling. Rippel and Bourdev's method~\cite{Rippel2017real} contains an autoencoder featuring pyramidal analysis and supplement their approach with adversarial training. They achieve real-time compression with pleasing visual quality at a rather low bit rate. 

In general, these existing DL-based compression codecs, like the conventional codecs, also only compress the images at pixel level, and do not consider the semantics of them.
%
%
%
%
%


\begin{figure*}[t]
 \centering
 \begin{minipage}[t]{0.95\linewidth}
 \centering
 \includegraphics[width=0.95\linewidth]{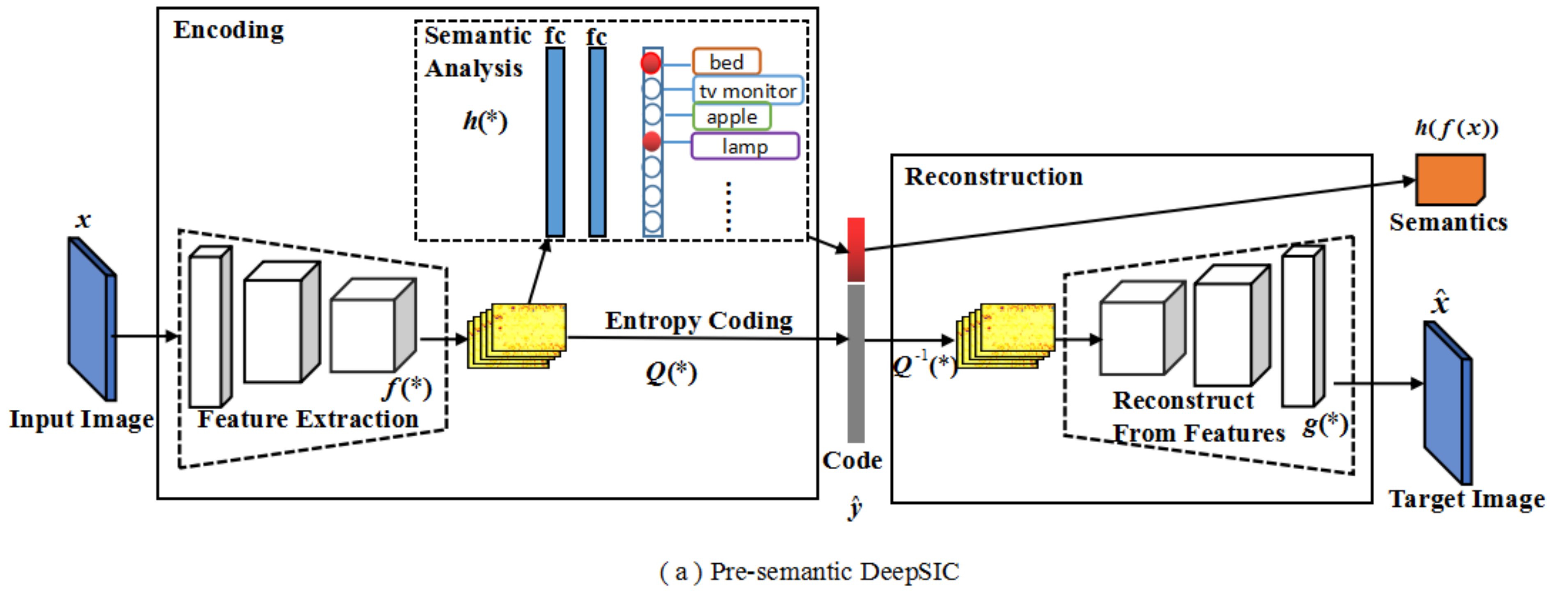}
 \end{minipage}%
 
 \begin{minipage}[t]{0.95\linewidth}
 \centering
 \includegraphics[width=0.95\linewidth]{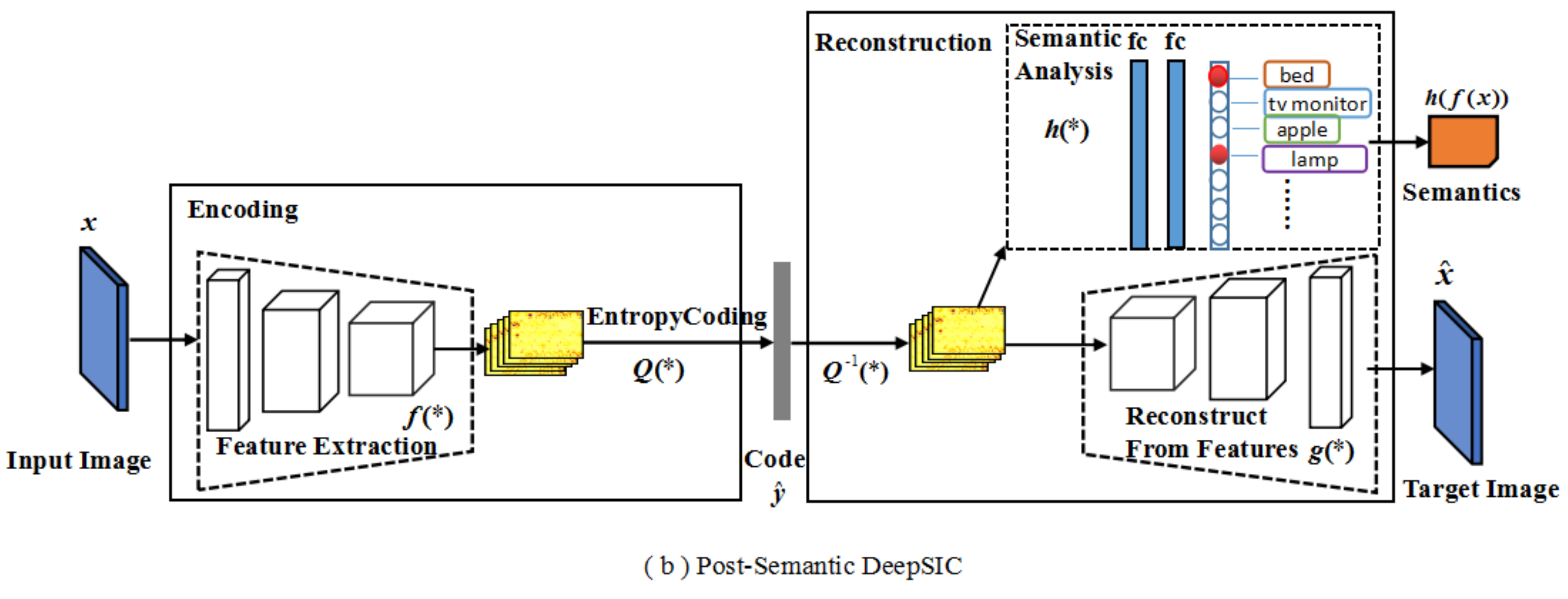}
 \end{minipage}%
   \caption{Two architectures of our semantic compression network: 
   (a) Image compression with pre semantic analysis in the encoder; 
   (b) Image compression with post semantic analysis in the decoder.}
\label{fig:twoarchitecture}
\end{figure*}

\section{Deep Semantic Image Compression}\label{sec3}

In the proposed deep semantic image compression framework~(DeepSIC), the compression scheme is similar to autoencoder. The encoder-decoder image compression pipeline commonly maps the target image through a bitrate bottleneck with an autoencoder and train the model to minimize a loss function penalizing it from its reconstruction result. For our DeepSIC, this requires a careful construction of a feature extractor and reconstruction module for the encoder and decoder, a good selection of an appropriate optimization objective for the joint-optimization of semantic analysis and compression, and an entropy coding module to further compress the fixed-sized feature map to gain codes with variable lengths.

Figure~\ref{fig:twoarchitecture} shows the two proposed DeepSIC architectures: pre-semantic DeepSIC and post-semantic DeepSIC respectively. For pre-semantic DeepSIC, it places the semantic analysis in the encoding process, which is implemented by reserving a portion of the bits in the compressed code to store semantic information. Hence the code innately reflects the semantic information of the image. For post-semantic image compression, the feature retained from decoding is used for the semantic analysis module to get the class label and for reconstruction network to synthesize the target image. Both architectures have modules of feature extraction, entropy coding, reconstruction from features and semantic analysis. 

Here, a brief introduction of the modules in our DeepSIC is given.\vspace{0.5em} 


\begin{figure*}[t]
\begin{center}
   \includegraphics[width=0.95\textwidth]{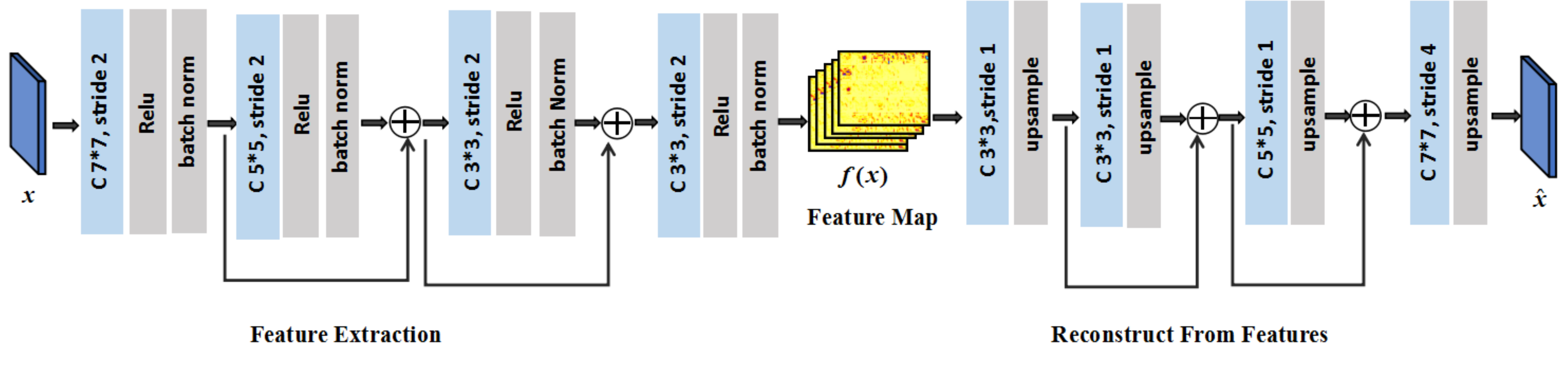}
\end{center}
   \caption{
   The feature extraction module and the reconstruction from features module: They are both formed by a four-stage convolutional network.}
\label{fig:featureExtract}
\end{figure*}

{\bf Feature Extraction}: Features represent different types of structure in images across scales and input channels. Feature extraction module in image compression aims to reduce the redundancy while maintaining max-entropy of the containing information. We adopt Convolutional Neural Network (CNN) as the feature extraction network. Given $x$ as the input image, $f(x)$ denotes the feature extraction output. 

{\bf Entropy Coding}: The feature output is firstly quantized to a lower bit precision. We then apply CABAC~\cite{Marpe2003Context} encoding method to lossless leverage the redundancy remained in the data. It encodes the output of quantization $y$ into the final binary code sequence \begin{math}\hat{y}\end{math}. 

{\bf Reconstruction from Features}: By entropy decoding, we retrieve image features from the compressed code. The inverse process of feature extraction is performed in the decoding process to reconstruct the target image.

{\bf Semantic Analysis}: Semantic analysis module is the analysis implemented on the extracted feature map. As mentioned in section \ref{sec:introduction}, there are many forms of semantic analysis such as semantic segmentation, object recognition, and image captioning. We adopt object classification as the semantic analysis module in this paper.\vspace{0.5em} 

%
%
%
%
%

In general, we encode the input images through feature extraction, and then quantize and code them into binary codes for storing or transmitting to the decoder. The reconstruction network then creates an estimate of the original input image based on the received binary code. We further train the network with semantic analysis both in the encoder and in the decoder (reconstruction) network. This procedure is repeated under the loss of distortion between the original image and the reconstruction target image, together with the error rate of the semantic analysis. The specific descriptions of each module are present in the subsequent subsections. 

\subsection{Feature Extraction}
The output of the feature extraction module is the feature map of an image, which contains the significant structure of the image. CNNs that create short paths from early layers to later layers allow feature reuse throughout the network, and thus allow the training of very deep networks. These CNNs can represent the image better and are demonstrated to be good feature extractors for various computer vision tasks~\cite{simonyan2014very, Long2015Fully, szegedy2015going, He2016Deep, huang2016densely,tong2017Image}. The strategy of reusing the feature throughout the network helps the training of deeper network architectures. Feature extraction model in compression model\cite{Theis2017lossy} adopt similar strategy. We adopt the operation of adding the batch-normalized output of the previous layer to the subsequent layer in the feature networks to . Furthermore, we also observe that the dense connections have a regularizing effect, which reduces overfitting on tasks with small training set sizes.

Our model extracts image features through a convolutional network illustrated in Figure~\ref{fig:featureExtract}. Given the input image as \begin{math} x\in \mathbb{R}^{C\times H \times W}\end{math}, we denote the feature extraction output as \begin{math}f(x)\end{math}.   

Specifically, our feature extraction module consists of four stages, each stage comprises a convolution, a subsampling, and a batch normalization layer. Each subsequent stage utilizes the output of the previous layers. And each stage begins with an affine convolution to increase the receptive field of the image. This is followed by \begin{math}4\times4\end{math}, \begin{math}2\times2\end{math}, or \begin{math}1\times1\end{math} downsampling to reduce the information. Each stage then concludes by a batch normalization operation.

\subsection{Entropy Coding}
Given the extracted tensor \begin{math}f(x)\in{R^{C\times H \times W}}\end{math}, before entropy coding the tensor, we first perform quantization. The feature tensor is optimally quantized to a lower bit precision $B$:

\begin{eqnarray}
Q(f(x))= \frac{1}{2^{B-1}} \left \lceil 2^{B-1}f(x) \right \rceil.
\end{eqnarray}

The quantization bin $B$ we use here is 6 bit. After quantization, the output is converted to a binary tensor. The entropy of the binary code generated during feature extraction and quantization period are not maximum because the network is not explicitly designed to maximize entropy in its code, and the model does not necessarily exploit visual redundancy over a large spatial extent. 

We exploit this low entropy by lossless compression via entropy coding, to be specific, we implement an entropy coding based on the context-adaptive binary arithmetic coding (CABAC) framework proposed by~\cite{Marpe2003Context}. Arithmetic entropy codes are designed to compress discrete-valued data to bit rates closely approaching the entropy of the representation, assuming that the probability model used to design the code approximates the data well. We associate each bit location in \begin{math}Q(f(x))\end{math} with a context, which comprises a set of features indicating the bit value. These features are based on the position of the bit as well as the values of neighboring bits. We train a classifier to predict the value of each bit from its context feature, and then use the resulting belief distribution to compress \begin{math}b\end{math}. 
 
Given \begin{math}y=Q(f(x))\end{math} denotes the quantized code, after entropy encoding \begin{math}y\end{math} into its binary representation \begin{math}\hat{y}\end{math}, we retrieve the compression code sequence.

During decoding, we decompress the code by performing the inverse operation. Namely, we interleave between computing the context of a particular bit using the values of previously decoded bits. The obtained context is employed to retrieve the activation probability of the bit to be decoded. Note that this constrains the context of each bit to only involve features composed of bits already decoded.

\subsection{Reconstruction From Features}

The module of reconstruction from features mirrors the structure of the feature extraction module, which is four-stage formed as well. Each stage comprises a convolutional layer and an upsampling layer. The output of each previous layer is passed on to the subsequent layer through two paths, one is the deconvolutional network, and the other is a straightforward upsampling to target size through interpolation. After reconstruction, we obtain the output decompressed image \begin{math}\hat{x}\end{math}.  

\begin{eqnarray}
\hat{x} = g\left({Q}^{-1}(Q(f(x)))\right)
\end{eqnarray}

Although arithmetic entropy encoding is lossless, the quantization will bring in some loss in accuracy, the result of \begin{math} {Q}^{-1}(Q(f(x))\end{math} is not exactly the same as the output of feature extraction. It is an approximation of \begin{math} f(x)\end{math}.

\subsection{ Semantic Analysis}
\begin{figure}[t]
\begin{center}
   \includegraphics[width=0.4\textwidth]{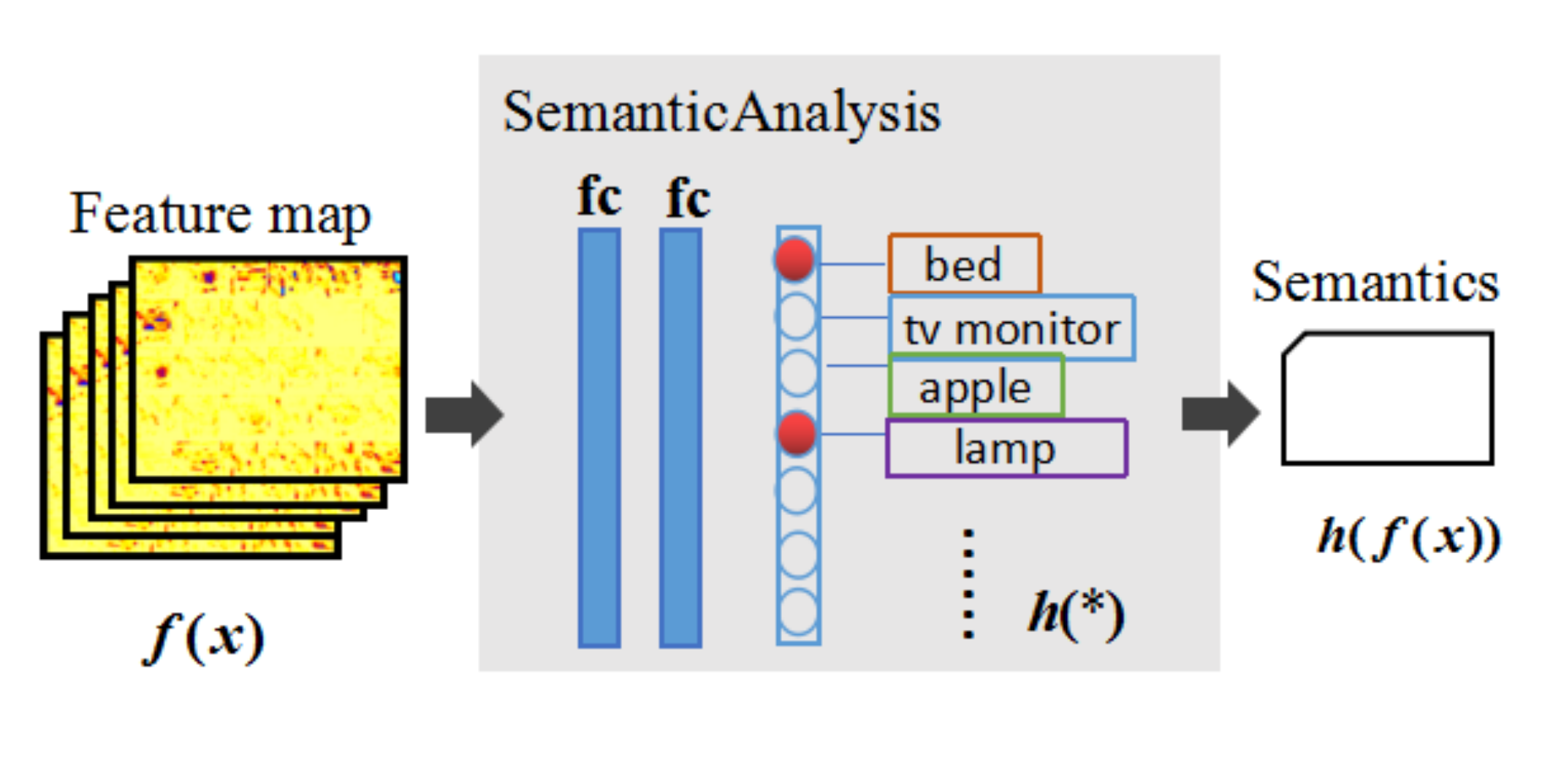}
\end{center}
   \caption{
   The structure of semantic analysis module}
\label{fig:semantic}
\end{figure}

As aforementioned, there are a number of semantic analysis forms. Classification task is the commonly selected way to evaluate deep learning networks~\cite{simonyan2014very, Long2015Fully, szegedy2015going, He2016Deep, huang2016densely}. Thus we select object classification for experiments in this paper. The structure of our semantic analysis module contains a sequence of convolutions following with two fully connected layers and a softmax layer.

Figure~\ref{fig:semantic} presents the structure of our semantic analysis module. It is position-optional and can be placed in the encoding and decoding process for the two different architectures. We denote it as \begin{math}h \left ( * \right )\end{math} to operate on the extracted feature map \begin{math}f\left ( x \right )\end{math}. Thus the output semantic representations are \begin{math}h \left ( f( x ) \right )\end{math}. 

For the classification task in the semantic analysis part, we adjust the learning rate using the related name-value pair arguments when creating the fully connected layer. Moreover, a softmax function is ultilized as the output unit activation function after the last fully connected layer for the multi-class classification. 

We set the cross entropy of the classification results as the semantic analysis loss \begin{math}L_{sem}\end{math} in this module. Denote the weight matrix of the two fully connected layer as \begin{math} W_{fc1}\end{math} and \begin{math} W_{fc2}\end{math} respectively. \begin{math}L_{sem}\end{math} is calculated as follows:

\begin{eqnarray}
L_{sem}=\mathbb{E}\left[ softmax\left[ W_{fc2}* \left( W_{fc1}* f(x) \right)\right]\right]
\end{eqnarray}

It is worth noting that the inputs of the semantic analysis module in the two proposed architectures are slightly different. The input feature maps of semantic analysis module in pre-semantic DeepSIC are under floating point precision. Differently, the input feature maps of semantic analysis module in post-semantic DeepSIC are under fixed-point precision due to quantization and entropy coding.

\subsection{Joint Training of Compression and Semantic Analysis}
We implement end-to-end training for the proposed DeepSIC, jointly optimize the two constraints of the semantic analysis and image reconstruction modules. And we define the loss as the weighted evaluation of compression ratio \begin{math}R\end{math}, distortion \begin{math}D\end{math} and the loss of the semantic analysis \begin{math}L_{sem}\end{math} in Equation \ref{equ:total_loss}. 

\begin{eqnarray} \label{equ:total_loss}　
L=R+{\lambda}_{1}D+{\lambda}_{2}L_{sem}
\end{eqnarray}

Here, \begin{math}$${\lambda}_{1}\end{math} and \begin{math}$$  {\lambda}_{2}\end{math} govern the trade-offs of the three terms. Since the quantization operation is non-differential, the marginal density of \begin{math}\hat{y_i}\end{math} is then given by the training of the discrete probability masses with weights equal to the probability mass function of \begin{math}\hat{y_i}\end{math}, where index $i$ runs over all elements of the vectors, including channels, image width and height.

\begin{eqnarray}
P_{y_i}\left ( n \right ) = \int_{n-\frac{1}{2}}^{n+\frac{1}{2}}p_{\hat{y_i}}(t)d_t
\end{eqnarray}
Therefore, \begin{math}R\end{math} can be calculated as

\begin{eqnarray}
R= \mathbb{E}\left [ \sum _i \log_2P_{y_i}\left ( n \right )  \right ].
\end{eqnarray}

\begin{math}$$D\end{math} measures the distortion introduced by coding and decoding. It's calculated by the distance between the original image and the reconstructed image. We take MSE as the distance metric for training, thus $D$ is defined as 

\begin{eqnarray}
D= \mathbb{E}\left [  {\left \| x_{i} - \hat{x}_{i} \right \|}_2^2 \right ].
\end{eqnarray}

\section{Experiment}\label{sec4}

In this section, we present experimental results over multiple datasets to demonstrate the effectiveness of the proposed semantic image compression. 

\subsection{Experimental Setup}
\noindent{\bf Datasets}  For training, we jointly optimized the full set of parameters over ILSVRC 2012 which is the subset of the ImageNet. The ILSVRC 2012 classification dataset consists of 1.2 million images for training, and 50,000 for validation from 1, 000 classes. A data augmentation scheme is adopted for training images and resize them to \begin{math}128\times128\end{math} at training time. We report the classification accuracy on the validation set. Performance tests on Kodak PhotoCD dataset are also present to enable comparison with other image compression codecs. Kodak PhotoCD dataset \cite{franzen1999kodak} is an uncompressed set of images which is popularly used for testing compression performances. 

\vspace{1em} 
\noindent{\bf Metric}  To assess the visual quality of reconstructed images, we adopt Multi-Scale Structural Similarity Index Metric~(MS-SSIM)~\cite{Wang2003Multiscale} and Peak Signal-to-Noise Ratio~(PSNR)~\cite{Gupta2012A} for comparing original, uncompressed images to compressed, degraded ones. We train our model on the MSE metric and evaluate all reconstruction model on MS-SSIM. MS-SSIM is a representative perceptual metric which has been specifically designed to match the human visual system. The distortion of reconstructed images quantified by PSNR is also provided to compare our method with JPEG, JPEG2000 and DL-based methods. Moreover, we report the classification accuracy over validation and testing sets to evaluate the performance of semantic analysis module.

\begin{figure}[t]
\begin{center}
   \includegraphics[width=0.48\textwidth]{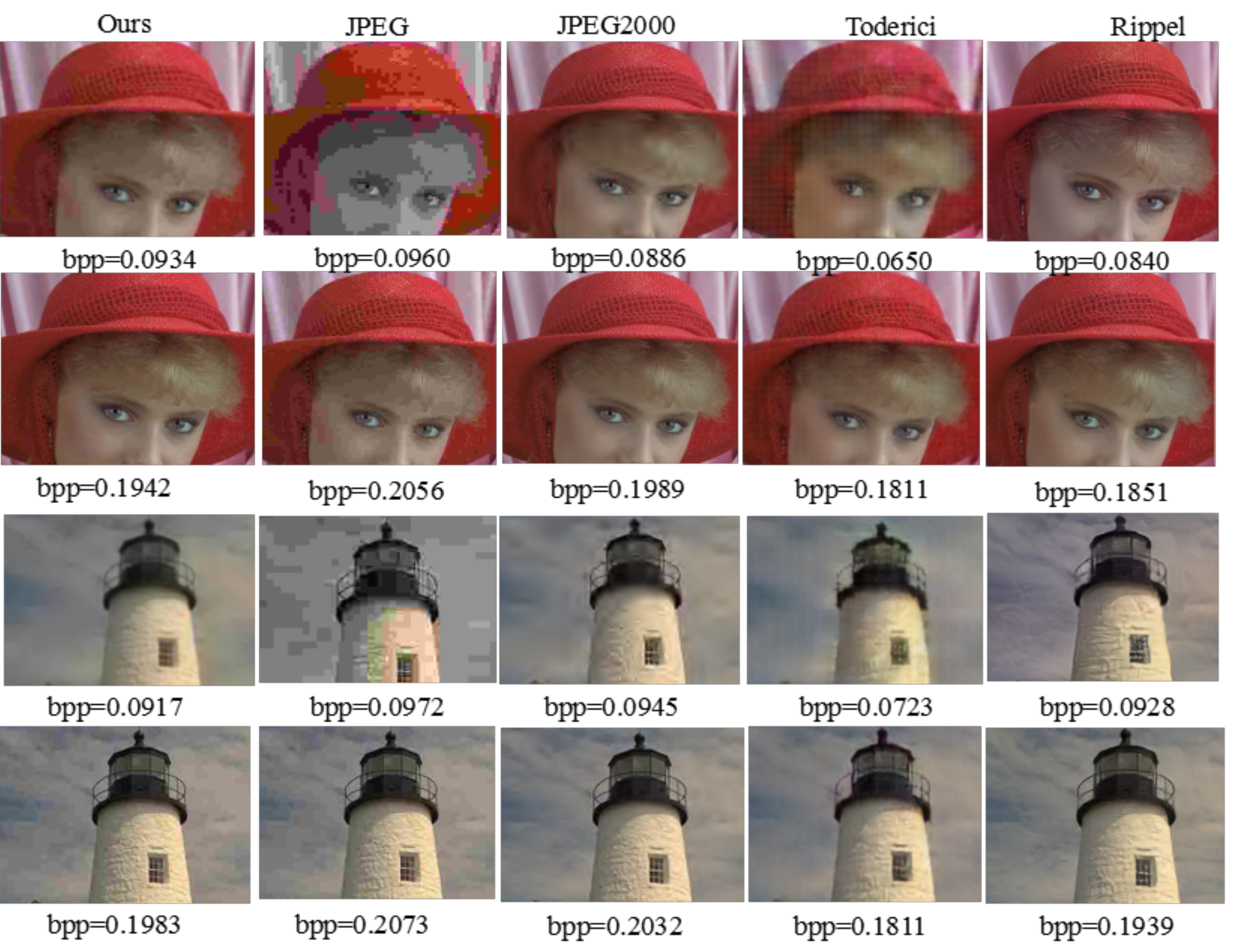}
\end{center}
   \caption{ Examples of reconstructed image parts by different codecs (JPEG, JPEG 2000, ours, Toderici~\cite{Toderici2016full} and Rippel~\cite{Rippel2017real}) for very low bits per pixel~(bpp) values. The uncompressed size is 24 bpp, so the examples represent compression by around 120 and 250 times. The test images are from the Kodak PhotoCD dataset.}
\label{fig:KodakResults}
\end{figure}

\begin{figure}[t]
\begin{center}
   \includegraphics[width=0.4\textwidth]{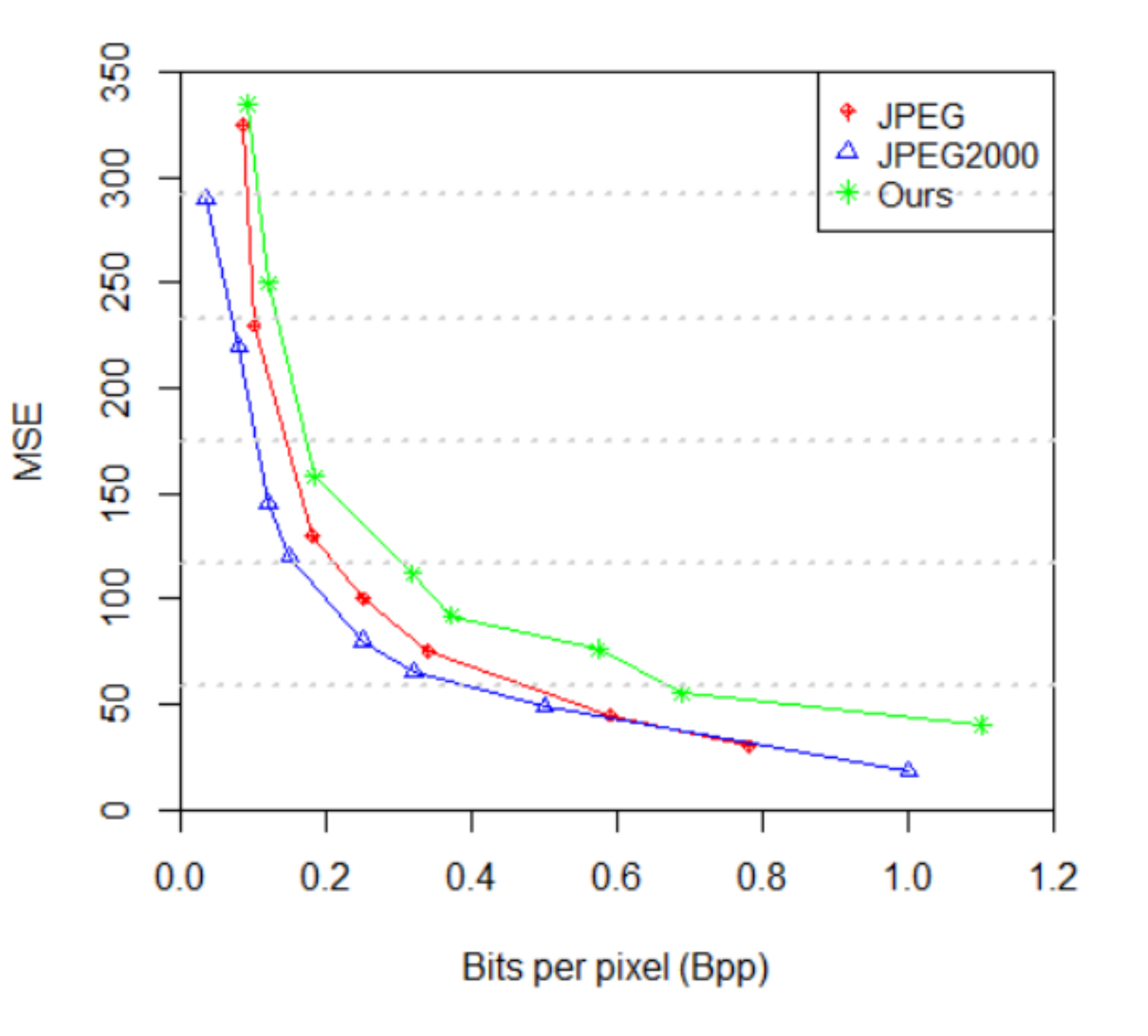}
\end{center}
   \caption{ Summary rate-distortion curves, computed by averaging results over the 24 images in the Kodak test set. JPEG and JPEG 2000 results are averaged over images compressed with identical quality settings.
   }
\label{fig:RDResults}
\end{figure}

%
%
%
%
\vspace{1em}

\subsection{Implementation and Training Details}
\begin{figure*}[htbp]
\begin{center}
   \includegraphics[width=0.95\textwidth]{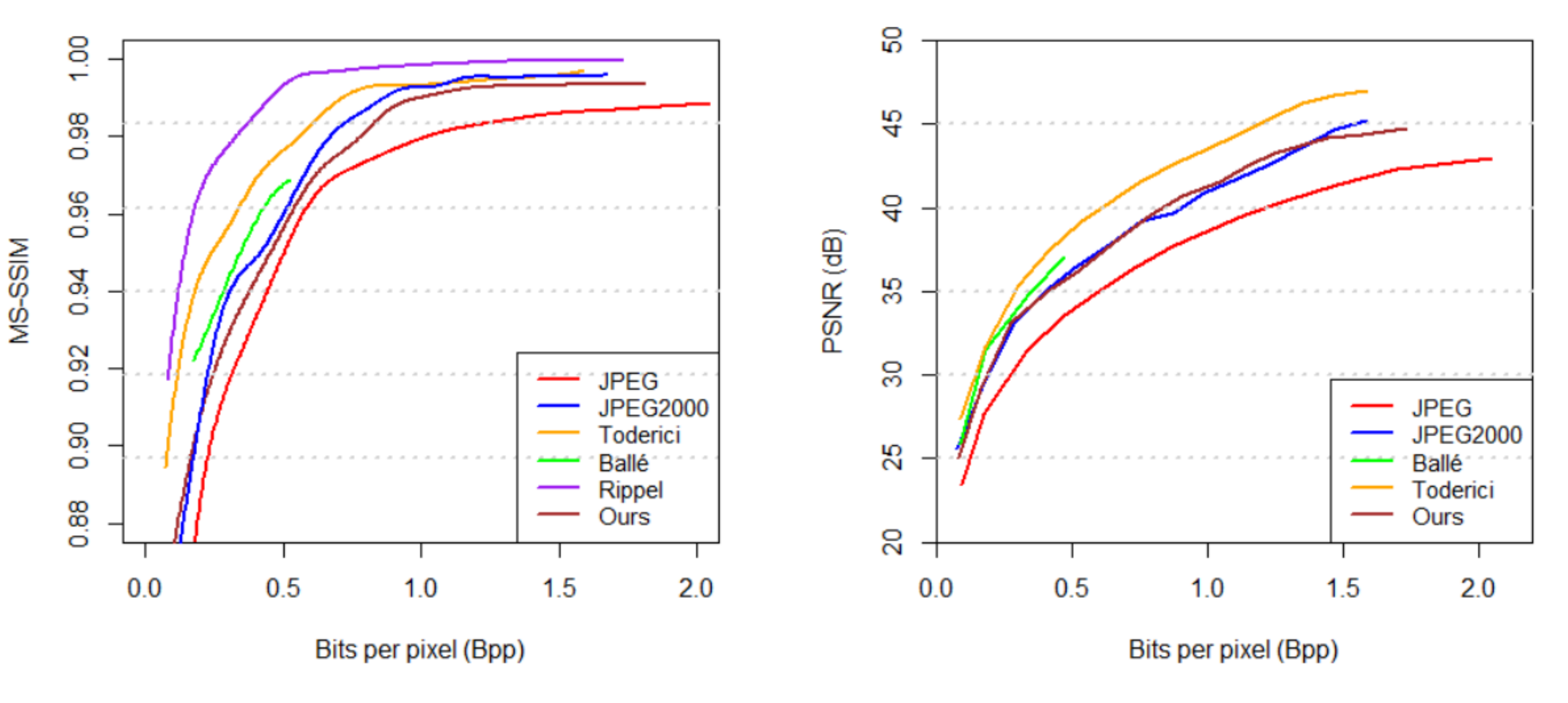}
\end{center}
   \caption{
   Average perceptual quality and compression ratio curves for the luma component of the images from Kodak dataset. Our DeepSIC is compared with JPEG, JPEG2000, Ball{\'e}~\cite{balle2016end}, Toderici~\cite{Toderici2016full} and Rippel~\cite{Rippel2017real}. Left: perceptual quality, measured with multi-scale structural similarity (MS-SSIM). Right: peak signal-to-noise ratio (PSNR). We have no access to reconstructions by Rippel\cite{Rippel2017real}, so we carefully transcribe their results, only available in MS-SSIM, from the graphs in their paper. 
   }
\label{fig:MSSSIMCurve}
\end{figure*}

We conduct training and testing on a NVIDIA Quadro M6000 GPU. All models are trained with 128\begin{math}$$\times\end{math}128 patches sampled from the ILSVRC 2012 dataset. All network architectures are trained using the Tensorflow API, with the Adam~\cite{kingma2014adam} optimizer. We set \begin{math}B=32\end{math} as the batch size, and \begin{math}C=3\end{math} as the number of color channels. The extracted feature dimensions is variable due to different subsample settings to gain variable length of compressed code. This optimization is performed separately for each weight, yielding separate transforms and marginal probability models. 
 
We use 128 filters~(size 7\begin{math}\times\end{math}7) in the first stage, each subsampled by a factor of 4 or 2 vertically and horizontally, and followed up with 128 filters (size 5\begin{math}\times\end{math}5) with the stride of 2 or 1. The remaining two stages retain the number of channels, but use filters operating across all input channels (\begin{math}3\times3\times128\end{math}), with outputs subsampled by a factor of 2 or 1 in each dimension. The structure of the reconstruction module mirrors the structure of the feature extraction module. 
%
%
%
%

The initial learning rate is set as 0.003, with decaying twice by a factor of 5 during training. We train each model for a total of 8,000,000 iterations. 

\subsection{Experimental Results}

Since semantic compression is a new concept, there are no direct baseline comparisons. We conduct many experiments and present their results as follows:

To evaluate performance of image compression quality, we compare DeepSIC against standard commercial compression techniques JPEG, JPEG2000, as well as recent deep-learning based compression work~\cite{balle2016end,Toderici2016full,Rippel2017real}. We show results for images in the test dataset and in every selected available compression rate. Figure~\ref{fig:KodakResults} shows visual examples of some images compressed using proposed DeepSIC optimized for a low value above 0.25 bpp, compared to JPEG, JPEG 2000 and DL-based images compressed at equal or greater bit rates. The average Rate-Distortion curves for the luma component of images in the Kodak PhotoCD dataset, shown in Figure~\ref{fig:RDResults}. Additionally, we take average MS-SSIM and PSNR over Kodak PhotoCD dataset as the functions of the bpp fixed for the testing images, shown in Figure~\ref{fig:MSSSIMCurve}. 

\begin{figure*}[t]
\begin{center}
   \includegraphics[width=0.98\textwidth]{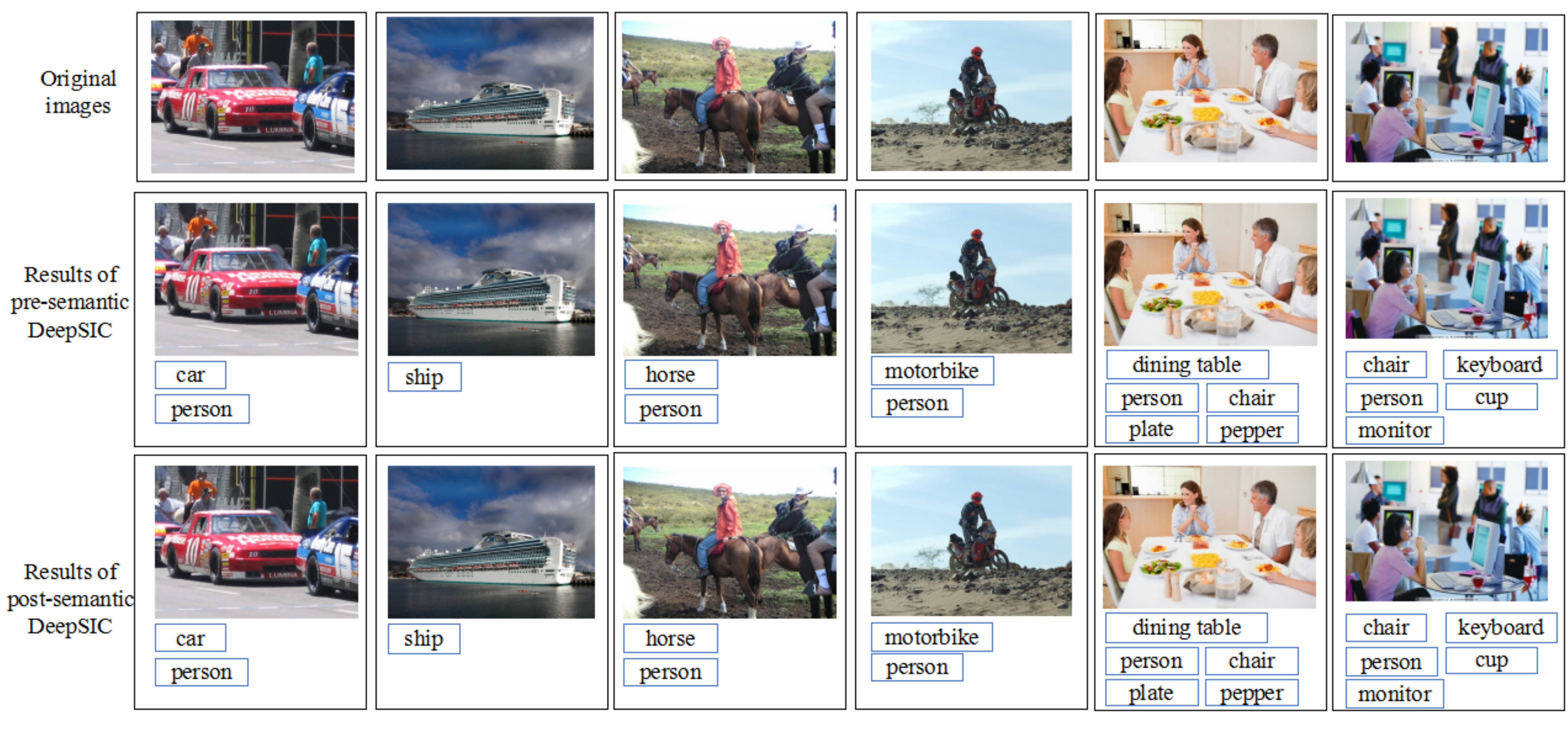}
\end{center}
   \caption{
   Examples of results with our DeepSIC: Images in the first row are the original images. The reconstructed images of the two architectures of DeepSIC are shown in the second and the third row followed by the output representation of the semantic analysis module. }
\label{fig:semanticResults}
\end{figure*}

To evaluate the performance of semantic analysis, we demonstrate some examples of the results of DeepSIC with reconstructions and their semantics, shown in Figure~\ref{fig:semanticResults}. 
Comparisons of the semantic analysis result of the two proposed architectures on classification accuracy are given in Table~\ref{tab:performance_comparison}. Furthermore, as different compression ratios directly affect the performance of compression, we compare semantic analysis result of the proposed architectures over certain fixed compression ratios with the mainstream classification methods in Table~\ref{tab:performance_comparison}. It also presents the trend of how compression ratio affects the performance of semantic analysis.

%
%
%
%

\begin{table}   
\begin{center}
\caption{Accuracy over different compression ratios (measured by bpp) on ILSVRC validation, with comparisons to the state-of-the-art classification methods. Pre-SA is short for pre-semantic DeepSIC and Post-SA is short for post-semantic DeepSIC.} 
\label{tab:performance_comparison} 
   \begin{tabular}{|l|c|c|}  
   \hline  
    Method & Top-1 acc. & Top-5 acc. \cr  
   \hline  
   \hline  
       Pre-SA (0.25bpp)  &52.2\%&72.7\%\cr 
       Post-SA (0.25bpp)  &51.6\%&71.4\%\cr\hline
       Pre-SA (0.5bpp)   &63.2\%&82.2\%\cr 
       Post-SA (0.5bpp)   &61.9\%&81.4\%\cr\hline
       Pre-SA (1.0bpp)   &68.7\%&89.4\%\cr
       Post-SA (1.0bpp)   &68.8\%&89.9\%\cr\hline
       Pre-SA (1.5bpp)   &67.1\%&90.1\%\cr
       Post-SA (1.5bpp)   &68.9\%&89.9\%\cr\hline
       VGG-16\cite{simonyan2014very} (10-crops) &71.9\%&90.7\%\cr
       Yolo Darknet\cite{redmon2016yolo9000} &76.5\%&93.3\%\cr
       DenseNet-121\cite{huang2016densely} (10-crops)   &76.4\%&93.4\%\cr
   \hline
\end{tabular}
\end{center}                
\end{table} 

Although we use MSE as a distortion metric for training and incorporate semantic analysis with compression, the appearance of compressed images are substantially comparable with JPEG and JPEG 2000, and slightly inferior to the DL-based image compression methods. Consistent with the appearance of these example images, we retain the semantic representation of them through the compression procedure. Although the performance of our method is neither the best on the visual quality of the reconstructed images nor on the classification accuracy, the result is still comparable with the state-of-the-art methods. 

\subsection{ Discussion}
We perform object classification as the semantic analysis module and represent the semantics with the identifier code of the class in this paper. Nevertheless, the semantics itself is complicated. The length of compressed code directly affect the size of compressed code and is far from limitless. The problem of how to efficiently organize semantic representation of multiple objects need careful consideration and exploration. 

Yolo9000~\cite{redmon2016yolo9000} performed classification with WordTree, a hierarchical model of visual concepts that can merge different datasets together by mapping the classes in the dataset to synsets in the tree. It inspires us that models like WordTree can also be applied to hierarchically encoding the semantics. We can set up a variety of levels to represent the semantic information. For example, from the low-level semantic representation, you can know  ``there is a cat in the image''. While from the high-level one, you can know not only  ``there's a cat'' but also ``what kind of cat it is''. This kind of schemes are more efficient to represent the semantics of the images.  
%
%
%
%
%

\section{Conclusion and Future Work}\label{sec5}
In this paper, we propose an image compression scheme incorporating semantics, which we refer to as Deep Semantic Image Compression~(DeepSIC). The proposed DeepSIC aims to reconstruct the images and generate corresponding semantic representations at the same time. We put forward two novel architectures of it: pre-semantic DeepSIC and post-semantic DeepSIC. To validate our approach, we conduct experiments on Kodak PhotoCD and ILSVRC datasets and achieve promising results. We also compare the performance of the proposed two architectures of our DeepSIC. Though incorporating semantics, the proposed DeepSIC is still comparable with the state-of-the-art methods over the experimental results. 

This practice opens up a new thread of challenges and has the potential to immediately impact a wide range of applications such as semantic compression of surveillance streams for the future smart cities, and fast post-transmission semantic image retrieval in the Internet of Things (IoT) applications.

Despite the challenges to explore, deep semantic image compression is still an inspiring new direction which breaks through the boundary of multi-media and pattern recognition. Nevertheless, it's unrealistic to explore all the challenges at once. Instead, we mainly focus on the framework of deep semantic image compression in this paper. The proposed DeepSIC paves a promising research avenue that we plan to further explore other possible solutions to the aforementioned challenges.

{\small
\bibliographystyle{ieee}
\bibliography{sihuibib,yezhoubib}

\begin{thebibliography}{10}\itemsep=-1pt

\bibitem{VQA}
S.~Antol, A.~Agrawal, J.~Lu, M.~Mitchell, D.~Batra, C.~L. Zitnick, and
  D.~Parikh.
\newblock Vqa: Visual question answering.
\newblock In {\em International Conference on Computer Vision (ICCV)}, 2015.

\bibitem{balle2016end}
J.~Ball{\'e}, V.~Laparra, and E.~P. Simoncelli.
\newblock End-to-end optimized image compression.
\newblock {\em arXiv preprint arXiv:1611.01704}, 2016.

\bibitem{chen2014learning}
X.~Chen and C.~L. Zitnick.
\newblock Learning a recurrent visual representation for image caption
  generation.
\newblock {\em arXiv preprint arXiv:1411.5654}, 2014.

\bibitem{donahue2014long}
J.~Donahue, L.~A. Hendricks, S.~Guadarrama, M.~Rohrbach, S.~Venugopalan,
  K.~Saenko, and T.~Darrell.
\newblock Long-term recurrent convolutional networks for visual recognition and
  description.
\newblock {\em arXiv preprint arXiv:1411.4389}, 2014.

\bibitem{franzen1999kodak}
R.~Franzen.
\newblock Kodak lossless true color image suite.
\newblock {\em Source: http://r0k. us/graphics/kodak}, 4, 1999.

\bibitem{Goodfellow2014Generative}
I.~J. Goodfellow, J.~Pouget-Abadie, M.~Mirza, B.~Xu, D.~Warde-Farley, S.~Ozair,
  A.~Courville, and Y.~Bengio.
\newblock Generative adversarial nets.
\newblock In {\em International Conference on Neural Information Processing
  Systems}, pages 2672--2680, 2014.

\bibitem{gregor2016towards}
K.~Gregor, F.~Besse, D.~J. Rezende, I.~Danihelka, and D.~Wierstra.
\newblock Towards conceptual compression.
\newblock In {\em Advances In Neural Information Processing Systems}, pages
  3549--3557, 2016.

\bibitem{Gupta2012A}
P.~Gupta, P.~Srivastava, S.~Bhardwaj, and V.~Bhateja.
\newblock A modified psnr metric based on hvs for quality assessment of color
  images.
\newblock In {\em International Conference on Communication and Industrial
  Application}, pages 1--4, 2012.

\bibitem{He2016Deep}
K.~He, X.~Zhang, S.~Ren, and J.~Sun.
\newblock Deep residual learning for image recognition.
\newblock In {\em IEEE Conference on Computer Vision and Pattern Recognition},
  pages 770--778, 2016.

\bibitem{Herranz2016Scene}
L.~Herranz, S.~Jiang, and X.~Li.
\newblock Scene recognition with cnns: Objects, scales and dataset bias.
\newblock In {\em IEEE Conference on Computer Vision and Pattern Recognition},
  pages 571--579, 2016.

\bibitem{huang2016densely}
G.~Huang, Z.~Liu, K.~Q. Weinberger, and L.~van~der Maaten.
\newblock Densely connected convolutional networks.
\newblock {\em arXiv preprint arXiv:1608.06993}, 2016.

\bibitem{im2016generating}
D.~J. Im, C.~D. Kim, H.~Jiang, and R.~Memisevic.
\newblock Generating images with recurrent adversarial networks.
\newblock {\em arXiv preprint arXiv:1602.05110}, 2016.

\bibitem{johnson2016perceptual}
J.~Johnson, A.~Alahi, and L.~Fei-Fei.
\newblock Perceptual losses for real-time style transfer and super-resolution.
\newblock In {\em European Conference on Computer Vision}, pages 694--711.
  Springer, 2016.

\bibitem{karpathy2014deep}
A.~Karpathy and F.-F. Li.
\newblock Deep visual-semantic alignments for generating image descriptions.
\newblock {\em arXiv preprint arXiv:1412.2306}, 2014.

\bibitem{kingma2014adam}
D.~Kingma and J.~Ba.
\newblock Adam: A method for stochastic optimization.
\newblock {\em arXiv preprint arXiv:1412.6980}, 2014.

\bibitem{Krizhevsky2012Using}
A.~Krizhevsky and G.~E. Hinton.
\newblock Using very deep autoencoders for content-based image retrieval.
\newblock In {\em European Symposium on Artificial Neural Networks, Bruges,
  Belgium}, 2011.

\bibitem{krizhevsky2012imagenet}
A.~Krizhevsky, I.~Sutskever, and G.~E. Hinton.
\newblock Imagenet classification with deep convolutional neural networks.
\newblock In {\em Advances in neural information processing systems}, pages
  1097--1105, 2012.

\bibitem{Long2015Fully}
J.~Long, E.~Shelhamer, and T.~Darrell.
\newblock Fully convolutional networks for semantic segmentation.
\newblock In {\em IEEE Conference on Computer Vision and Pattern Recognition},
  pages 3431--3440, 2015.

\bibitem{lu2016hierarchical}
J.~Lu, J.~Yang, D.~Batra, and D.~Parikh.
\newblock Hierarchical question-image co-attention for visual question
  answering.
\newblock {\em arXiv preprint arXiv:1606.00061}, 2016.

\bibitem{Marpe2003Context}
D.~Marpe, H.~Schwarz, and T.~Wiegand.
\newblock Context-based adaptive binary arithmetic coding in the h. 264/avc
  video compression standard.
\newblock {\em IEEE Transactions on circuits and systems for video technology},
  13(7):620--636, 2003.

\bibitem{rabbani2002overview}
M.~Rabbani and R.~Joshi.
\newblock An overview of the jpeg 2000 still image compression standard.
\newblock {\em Signal processing: Image communication}, 17(1):3--48, 2002.

\bibitem{redmon2016yolo9000}
J.~Redmon and A.~Farhadi.
\newblock Yolo9000: better, faster, stronger.
\newblock {\em arXiv preprint arXiv:1612.08242}, 2016.

\bibitem{Rippel2017real}
O.~Rippel and L.~Bourdev.
\newblock Real-time adaptive image compression.
\newblock {\em arXiv preprint arXiv:1705.05823}, 2017.

\bibitem{simonyan2014very}
K.~Simonyan and A.~Zisserman.
\newblock Very deep convolutional networks for large-scale image recognition.
\newblock {\em arXiv preprint arXiv:1409.1556}, 2014.

\bibitem{szegedy2015going}
C.~Szegedy, W.~Liu, Y.~Jia, P.~Sermanet, S.~Reed, D.~Anguelov, D.~Erhan,
  V.~Vanhoucke, and A.~Rabinovich.
\newblock Going deeper with convolutions.
\newblock In {\em IEEE Conference on computer vision and pattern recognition},
  pages 1--9, 2015.

\bibitem{Theis2017lossy}
L.~Theis, W.~Shi, A.~Cunningham, and F.~Husz{\'a}r.
\newblock Lossy image compression with compressive autoencoders.
\newblock {\em arXiv preprint arXiv:1703.00395}, 2017.

\bibitem{Toderici2015variable}
G.~Toderici, S.~M. O'Malley, S.~J. Hwang, D.~Vincent, D.~Minnen, S.~Baluja,
  M.~Covell, and R.~Sukthankar.
\newblock Variable rate image compression with recurrent neural networks.
\newblock {\em arXiv preprint arXiv:1511.06085}, 2015.

\bibitem{Toderici2016full}
G.~Toderici, D.~Vincent, N.~Johnston, S.~J. Hwang, D.~Minnen, J.~Shor, and
  M.~Covell.
\newblock Full resolution image compression with recurrent neural networks.
\newblock 2017.

\bibitem{tong2017Image}
L.~G. L.~X. Tong, T. and Q.~Gao.
\newblock Image super-resolution using dense skip connections.
\newblock In {\em IEEE Conference on Computer Vision and Pattern Recognition}.

\bibitem{Van2010Evaluating}
K.~E.~A. Van, de~Sande, T.~Gevers, and C.~G.~M. Snoek.
\newblock Evaluating color descriptors for object and scene recognition.
\newblock {\em IEEE Transactions on Pattern Analysis and Machine Intelligence},
  32(9):1582--96, 2010.

\bibitem{vinyals2014show}
O.~Vinyals, A.~Toshev, S.~Bengio, and D.~Erhan.
\newblock Show and tell: A neural image caption generator.
\newblock {\em arXiv preprint arXiv:1411.4555}, 2014.

\bibitem{Wallace1992The}
G.~K. Wallace.
\newblock The jpeg still picture compression standard.
\newblock {\em Communications of the Acm}, 38(1):xviii--xxxiv, 1992.

\bibitem{Wang2011Action}
H.~Wang, A.~Kläser, C.~Schmid, and C.~L. Liu.
\newblock Action recognition by dense trajectories.
\newblock In {\em IEEE Conference on Computer Vision and Pattern Recognition},
  pages 3169--3176, 2011.

\bibitem{Wang2003Multiscale}
Z.~Wang, E.~P. Simoncelli, and A.~C. Bovik.
\newblock Multiscale structural similarity for image quality assessment.
\newblock In {\em Signals, Systems and Computers, 2004. Conference Record of
  the Thirty-Seventh Asilomar Conference on}, pages 1398--1402 Vol.2, 2003.

\bibitem{NIPS2014_5349}
B.~Zhou, A.~Lapedriza, J.~Xiao, A.~Torralba, and A.~Oliva.
\newblock Learning deep features for scene recognition using places database.
\newblock In Z.~Ghahramani, M.~Welling, C.~Cortes, N.~D. Lawrence, and K.~Q.
  Weinberger, editors, {\em Advances in Neural Information Processing Systems
  27}, pages 487--495. Curran Associates, Inc., 2014.

\bibitem{zhu2015visual7w}
Y.~Zhu, O.~Groth, M.~Bernstein, and L.~Fei-Fei.
\newblock Visual7w: Grounded question answering in images.
\newblock {\em arXiv preprint arXiv:1511.03416}, 2015.

\end{thebibliography}
}

\end{document}